\documentclass{ecai}

\usepackage{microtype}
\usepackage{bbm}
\usepackage{graphicx}
\usepackage{mathtools} 
\usepackage{subcaption}
\usepackage{booktabs} 

\usepackage{hyperref}

\newcommand{\CNF}{\textsf{CNF}}

\usepackage{amsmath}
\usepackage{amssymb}
\usepackage{mathtools}
\usepackage{amsthm}
\usepackage{hyperref}
\usepackage{url}
\usepackage{amsmath}
\usepackage{amssymb}
\usepackage{mathtools}
\usepackage{amsthm}
\usepackage{graphicx}
\usepackage{multirow}
\usepackage{microtype}
\usepackage{graphicx}
\usepackage{booktabs} 
\usepackage{graphicx}
\usepackage{capt-of}
\usepackage{caption}
\usepackage{booktabs}
\usepackage{varwidth}
\usepackage{svg}            
\usepackage{tikz}
\usepackage{comment}

\usepackage{amsmath,amssymb} 
\usepackage{color}
\usepackage{hyperref}   
\usepackage{booktabs}
\usepackage{multirow}
\usepackage{caption}
\usepackage{svg}  
\usepackage{wrapfig,lscape}
\usepackage{float}
\usepackage{rotating}
\usepackage{multicol}
\usepackage{listings}
\usepackage{tikz}
\usepackage{makecell}
\usepackage{comment}

\usepackage{pgfplots}
\usetikzlibrary{positioning, shapes, arrows, chains, intersections}
\newcommand\xstart{0}

\usetikzlibrary{calc}
\usetikzlibrary{decorations.pathreplacing}
\usetikzlibrary{tikzmark,decorations.pathreplacing,calligraphy}
\definecolor{jonquil}{rgb}{0.98, 0.85, 0.37}
\definecolor{babypink}{rgb}{0.96, 0.73, 0.73}
\definecolor{babypink2}{rgb}{0.96, 0.82, 0.82}
\definecolor{babyblueeyes}{rgb}{0.57, 0.73, 0.92}
\definecolor{babyblueeyes2}{rgb}{0.63, 0.79, 0.95}
\definecolor{darkseagreen}{rgb}{0.63, 0.83, 0.63}
\definecolor{darkseagreen2}{rgb}{0.63, 0.92, 0.63}

\usepackage[capitalize,noabbrev]{cleveref}

\theoremstyle{plain}

\theoremstyle{definition}

\theoremstyle{remark}

\usepackage{bm}

\usepackage[textsize=tiny]{todonotes}

\newlength{\subseclen}
\begin{document}

\begin{frontmatter}

\title{A New Interpretable Neural Network-Based Rule Model for Healthcare Decision Making}


\author[*]{\fnms{Adrien}~\snm{Benamira}}
\author[*]{\fnms{Tristan}~\snm{Guérand}}
\author{\fnms{Thomas}~\snm{Peyrin}} 

\address{Nanyang Technological University, Singapore}
\address[*]{Main contribution}


\begin{abstract}

In healthcare applications, understanding how machine/deep learning models make decisions is crucial. In this study, we introduce a neural network framework, \textit{Truth Table rules} (TT-rules), that combines the global and exact interpretability properties of rule-based models with the high performance of deep neural networks. TT-rules is built upon \textit{Truth Table nets} (TTnet), a family of deep neural networks initially developed for formal verification. By extracting the necessary and sufficient rules $\mathcal{R}$ from the trained TTnet model (global interpretability) to yield the same output as the TTnet (exact interpretability), TT-rules effectively transforms the neural network into a rule-based model. This rule-based model supports binary classification, multi-label classification, and regression tasks for small to large tabular datasets. 

After outlining the framework, we evaluate TT-rules' performance on healthcare applications and compare it to state-of-the-art rule-based methods. Our results demonstrate that TT-rules achieves equal or higher performance compared to other interpretable methods. Notably, TT-rules presents the first accurate rule-based model capable of fitting large tabular datasets, including two real-life DNA datasets with over 20K features.

\end{abstract}

\end{frontmatter}

\vspace{-10pt}

\section{Related Work}
Traditional rule-based models, such as decision trees \cite{bessiere2009minimising}, rule lists \cite{rivest1987learning, angelino2017learning, dash2018boolean}, linear models, and rule sets \cite{lakkaraju2016interpretable, cohen1995fast, cohen1999simple, quinlan2014c4, wei2019generalized}, are commonly used for interpretable classification and regression tasks. However, these models face limitations in handling large datasets, binary classification tasks, and capturing complex feature relationships, which can result in reduced accuracy and limited practicality \cite{yang2021learning, wang2021scalable}. To overcome these challenges, recent work by Benamira \textit{et al.} introduced an architecture encoded into $\CNF$ formulas, demonstrating scalability on large datasets \cite{10.1007/978-3-031-25056-9_31, biere2009handbook}. Our objective is to extend this approach to handle diverse classification tasks and regression on tabular datasets of varying feature dimensions.

There have been investigations into the connection between deep neural networks (DNNs) and rule-based models. Notable works include DNF-net \cite{abutbul2020dnf}, which focuses on the activation function, and RRL \cite{wang2021scalable}, which addresses classification tasks but raises concerns about interpretability due to its complexity and time-consuming training process. Another architecture, Neural Additive Models (NAMs) \cite{agarwal2020neural}, combines the flexibility of DNNs with the interpretability of additive models but deviates from the strict rule-based model paradigm, posing challenges in interpretation, especially with a large number of features. 

\vspace{-10pt}

\section{Methodology}

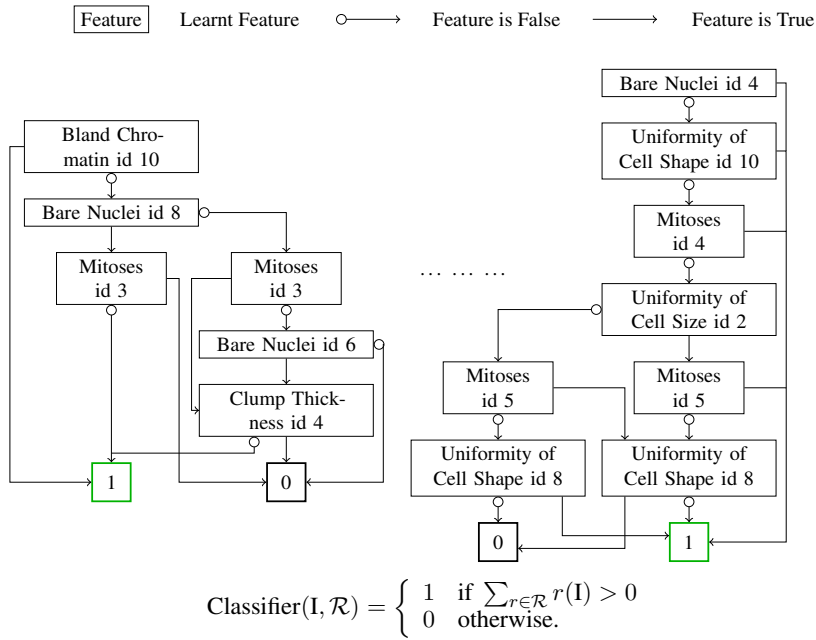
\begin{figure*}[htb!]
\centering
\captionof{figure}{\label{fig:cancer_casestudy} Our neural network model trained on the Breast Cancer dataset in the form of Boolean decision trees: the output of the DNN and the output of these decision trees are the same, reaching 99.30\% AUC. On the same test set, Random Forest reaches 95.08\% AUC, Decision Tree 90.36\% AUC and XGboost 97.79\% AUC. Each rule $r_i$ is a function $r_i : \{0,1\}^n \mapsto \{-1,0,1\}$, i.e for each data sample I we associate for each rule $r_i$ a score which is in $\{-1,0,1\}$. The prediction of our classifier is then as stated above. As our model has 24 rules, we have only reported two positive rules out of the 24 to provide an example of the type of rules obtained.
}
\resizebox{1.25\columnwidth}{!}{
\begin{tikzpicture}[every text node part/.style={align=center}, on top/.style={preaction={draw=white,-,line width=#1}},
on top/.default=4pt]

\tikzset{
coord/.style={coordinate, on chain, on grid, node distance=6mm and 25mm},
}
    \colorlet{darkgreen}{black!30!green};
    \colorlet{darkorange}{rgb:red,5;green,1};

    \node[rectangle,draw] (L1) at (\xstart,2)  {Feature};
    \node[rectangle] (L11) at (\xstart+2,2) {Learnt Feature};


    \node[coordinate] (L31) at (\xstart-0.5+2+2,2) {};
    \node[coordinate] (L32) at (\xstart+0.5+2+2,2) {};
    \draw[o->] (L31) -- (L32);
    \node[rectangle] (L33) at (\xstart+2+0.5+3.5,2) {Feature is False};

    \node[coordinate] (L41) at (\xstart-0.5+2+6,2) {};
    \node[coordinate] (L42) at (\xstart+0.5+2+6,2) {};
    \draw[->] (L41) -- (L42);
    \node[rectangle] (L43) at (\xstart+2+0.5+7.5,2) {Feature is True};

    \node[rectangle,draw, text width=2.5cm] (A1) at (\xstart,0)  {Bland Chromatin id 10};
    \node[rectangle,draw, below=0.4 of A1, text width=2.5cm] (B1) {Bare Nuclei id 8};
    
    \node[rectangle,draw, below=0.4 of B1, text width=1.5cm] (C1) {Mitoses id 3};
    \node[rectangle,draw, right=1cm of C1.east, text width=1.5cm] (D1) {Mitoses id 3};
    
    \node[rectangle,draw, below=0.4 of D1, text width=2.5cm] (F1) {Bare Nuclei id 6};
    \node[rectangle,draw, below=0.4 of F1, text width=2.5cm] (E1) {Clump Thickness id 4};

    \node[rectangle,draw,thick,minimum height=6mm,minimum width=6mm, below=0.4 of E1] (label11) {0};

    \node[coordinate, below of=C1, node distance=40] (c11) {}; 
    \node[coordinate, left of=label11, node distance=15] (c21) {}; 
    
    \node[rectangle,draw=darkgreen,thick, minimum height=6mm,minimum width=6mm] (label12) at (intersection of C1--c11 and label11--c21) {1};
    \node[coordinate, left of=label12, node distance=45] (c31) {};

    \node[coordinate, right of=C1, node distance=30] (c41) {};
    \node[coordinate, left of=D1, node distance=42] (c51) {};
    \node[coordinate, above of=label12, node distance=13] (c61) {};
    \node[coordinate, right of=label11, node distance=43] (c71) {};




    \draw[o->] (A1.south) -- (B1);
    \draw[->] (A1.west) -| (c31) -- (label12) ;

    \draw[->] (B1.south) -- (C1);
    \draw[o->] (B1.east) -| (D1);
    
    \draw[o->] (C1.south) -- (label12);
    \draw[->] (C1.east) -- (c41) |- (label11.west);

    \draw[o->] (D1.south) -- (F1);
    \draw[->] (D1.west) -- (c51) |- (E1.west);

    \draw[->] (E1.south) -- (label11);
    \draw[o-] ([xshift=-0.5cm] E1.south) |- (c61);

    \draw[->] (F1.south) -- (E1);
    \draw[o->] (F1.east) -| (c71) -- (label11);

    \node[rectangle,draw, text width=2.5cm] (A2) at (\xstart+9,1)  {Bare Nuclei id 4};
    \node[rectangle,draw, below=0.4 of A2, text width=2.5cm] (B2) {Uniformity of Cell Shape id 10};
    \node[rectangle,draw, below=0.4 of B2, text width=1.5cm] (C2) {Mitoses id 4};
    \node[rectangle,draw, below=0.4 of C2, text width=2.5cm] (D2) {Uniformity of Cell Size id 2};
    
    \node[rectangle,draw, below=0.4 of D2, text width=1.5cm] (E2) {Mitoses id 5};
    \node[rectangle,draw, left=1.25cm of E2.west, text width=1.5cm] (F2) {Mitoses id 5};
    \node[rectangle,draw, below=0.4 of E2, text width=2.5cm] (G2) {Uniformity of Cell Shape id 8};
    \node[rectangle,draw, below=0.4 of F2, text width=2.5cm] (H2) {Uniformity of Cell Shape id 8};

    \node[rectangle,draw,thick,minimum height=6mm,minimum width=6mm, below=0.4 of H2] (label21) {0};
    \node[rectangle,draw=darkgreen,thick, minimum height=6mm,minimum width=6mm, below=0.4 of G2] (label22) {1};

    \node[coordinate, right of=A2, node distance=43] (c12) {}; 
    \node[coordinate, right of=B2, node distance=43] (c22) {};
    \node[coordinate, right of=C2, node distance=43] (c32) {};
    \node[coordinate] (c42) at (intersection of c12--c22 and C2--c32) {};
    \node[coordinate, right of=E2, node distance=43] (c52) {};
    \node[coordinate] (c62) at (intersection of c12--c22 and E2--c52) {};

    \draw[o->] (A2.south) -- (B2);
    \draw[-] (A2.east) -- (c12) -- (c22) ;

    \draw[o->] (B2.south) -- (C2);
    \draw[-] (B2.east) -- (c22) -- (c42) ;

    \draw[o->] (C2.south) -- (D2);
    \draw[-] (C2.east) -- (c42) -- (c62);

    \draw[o->] (D2.west) -| (F2);
    \draw[->] (D2.south) -- (E2);
    
    \draw[->] (E2.east) -- (c62) |- (label22.east) ;
    \draw[o->] (E2.south) -- (G2);

    \draw[->] (F2.east) -| ([xshift=-1cm] G2.north);
    \draw[o->] (F2.south) -- (H2);

    \draw[->] ([xshift=-1cm] G2.south) |- ([yshift=-0.1cm] label21.east);
    \draw[o->] (G2.south) -- (label22);

    \draw[->] ([xshift=+1cm] H2.south) |- ([yshift=+0.1cm] label22.west);
    \draw[o->] (H2.south) -- (label21);


     \foreach \i in {1,2,3} {
    \node at (4.5 + \i*0.5, -2) {\dots};
  }

  \node[rectangle] (equ) at (\xstart+5, -7.25) {\large
$\text{Classifier}(\text{I}, \mathcal{R}) = \left\{
    \begin{array}{ll}
        1 & \mbox{if } \sum_{r \in \mathcal{R}} r(\text{I}) > 0 \\
        0 & \mbox{otherwise.}
    \end{array}
\right. 
$
};

    
\end{tikzpicture}
}

\end{figure*}

This paper introduces a novel neural network framework that effectively combines the interpretability of rule-based models with the high performance of DNNs. Our framework, called TT-rules, builds upon the advancements made by Benamira \textit{et al.}~\cite{10.1007/978-3-031-25056-9_31}
Benamira \textit{et al.}~\cite{10.1007/978-3-031-25056-9_31} introduced a new Convolutional Neural Network (CNN) filter function called the Learning Truth Table (LTT) block. The LTT block has the unique property of its complete distribution being computable in constant and practical time, regardless of the architecture. This allows the transformation of the LTT block from weights into an exact mathematical Boolean formula. Since an LTT block is equivalent to a CNN filter, \textbf{the entire neural network model, known as Truth Table net (TTnet), can itself be represented as a Boolean formula}. 
We then optimize our formula set $\mathcal{R}$ in two steps. 
We automatically integrate human logic into the truth tables. This reduces the size of each rule in the set $\mathcal{R}$. Then we analyze the correlation to decrease the number of rules in $\mathcal{R}$. These optimizations, specific to the TT-rules framework, automatically and efficiently transform the set $\mathcal{R}$ into an optimized set in constant time.
To enhance the interpretability of the model, we convert all boolean formulas into Reduced Ordered Binary Decision Diagrams. An example is given Figure~\ref{fig:cancer_casestudy}.



\begin{figure*}[!htb]
\center
\captionof{table}{\label{tab:perfs} Comparison machine learning dataset of our method to Linear/Logistic Regression~\cite{pedregosa2011scikit}, Decision Trees (DT)~\cite{pedregosa2011scikit}, GL~\cite{wei2019generalized}, 
Random Forest~\cite{ho1995random} and DNNs. Means and standard deviations are reported from 5-fold cross
validation. Our TT-rules models were trained with a final linear regression with weights as floating points for the Breast Cancer and Diabetes dataset for better performances. The two other datasets were trained with a sparse binary linear regression to reduce the number of final features. The lower RMSE the better, the higher AUC/Accuracy the better.}
\resizebox{1.25\columnwidth}{!}{
\begin{tabular}{@{}l|c|cc|c}
\toprule
 & \textbf{Regression} (RMSE)  & \multicolumn{2}{c|}{\textbf{Binary classification} (AUC)}  & \textbf{Multi-classification} (Accuracy) \\ \midrule
 & TCCA Cancer & Melanoma & Breast Cancer     & Diabetes                                             \\ 
 continous/binary \# & 0/20530 & 0/23689 features    & 0/81 features     & 43/296 features                                           \\ \midrule
Linear/ log          & 0.092  & 0.833 & 0.985 & 0.581                                               \\
DT       & -     & - & 0.908 & 0.572                                              \\
GL                   & -     & - & 0.984 & -                                                \\
TT-rules (Ours)                 & 0.029    & 0.835 & 0.986 & 0.584                                       \\ \hline 
Random Forest              & 0.42    & 0.729 & 0.950 & 0.587                                                  \\
DNNs                 & 0.028     & 0.725 & 0.982 & 0.603                                                    \\ \bottomrule
\end{tabular}}
\end{figure*}

\vspace{-10pt}
\section{Experiments}

\subsection{Datasets}

We utilized a variety of healthcare datasets for our study, including the Diabetes 130 US-Hospitals dataset for multi-classification\footnote{\href{https://bit.ly/diabetes_130_uci}{https://bit.ly/diabetes\_130\_uci}} \cite{Dua:2019}, two single-cell RNA-seq analysis datasets (head and neck cancer\footnote{\href{https://bit.ly/neck_head_rna}{https://bit.ly/neck\_head\_rna}} \cite{Puram2017SingleCell} and melanoma\footnote{\href{https://bit.ly/melanoma_rna}{https://bit.ly/melanoma\_rna}} \cite{Tirosh2016Dissecting}), the Breast Cancer Wisconsin (Original) dataset\footnote{\href{https://archive.ics.uci.edu/dataset/15/breast+cancer+wisconsin+original}{https://archive.ics.uci.edu/dataset/15/breast+cancer+wisconsin+original}}, and the TCGA lung cancer dataset for regression\footnote{\href{https://bit.ly/tcga_lung_rna}{https://bit.ly/tcga\_lung\_rna}} \cite{liu2018integrated}.

Our TT-rules framework's scalability is demonstrated using two DNA datasets. These include single-cell RNA-seq analysis datasets for head and neck cancer, melanoma cancer \cite{Puram2017SingleCell, Tirosh2016Dissecting}, and the TCGA lung cancer dataset \cite{liu2018integrated}. These datasets contain 23689 and 20530 features, respectively, and are commonly used in real-life machine learning applications \cite{li2017comprehensive,grewal2019application,revkov2023puree,tran2021fast}. In the melanoma cancer setup, we trained on the head and neck dataset \cite{Puram2017SingleCell} and tested on the melanoma dataset \cite{Tirosh2016Dissecting} following established literature \cite{li2017comprehensive,grewal2019application,revkov2023puree,tran2021fast}.

\vspace{-10pt}

\subsection{Performance Comparison}

Table~\ref{tab:perfs} presents a comparison of various rule-based models, including ours, on the datasets introduced before, in terms of RMSE, AUC and Accuracy. Our proposed model outperforms the others in terms of accuracy on the Diabetes dataset and on the Breast Cancer dataset. XGBoost and DNNs performs better on Diabetes but worse on bigger datasets as shown in the next section. Although GL provides a better tradeoff between performance and complexity, we highlight that GL does not support multi-class classification tasks and is not scalable for larger datasets such as DNA datasets, as shown in the next section.

\vspace{-10pt}

\subsection{Scalability Comparison}
\label{subsec:scalability}

Our TT-rules framework demonstrated excellent scalability to real-life datasets with up to 20K features. This result is not surprising, considering the original TTnet paper~\cite{10.1007/978-3-031-25056-9_31} showed the architecture's ability to scale to ImageNet. Furthermore, our framework's superiority was demonstrated by outperforming other rule-based models that failed to converge to such large datasets (GL~\cite{wei2019generalized}).  Regarding performance, the TT-rules framework outperforms all other methods.
Our approach not only scales but also reduces the input feature set, acting as a feature selection method. We generated a set of 1064 rules out of 20530 features for the regression problem, corresponding to a drastic reduction in complexity. For the binary classification dataset, we generated 9472 rules, which more then halved the input size from 23689 to 9472.

\vspace{-10pt}

\section{Conclusion}

In conclusion, our proposed TT-rules framework provides a new and optimized approach for achieving global and exact interpretability in regression and classification tasks. With its ability to scale on large datasets and its potential for feature reduction, the TT-rules framework appears as a valuable tool towards explainable artificial intelligence for healthcare applications.


\begin{thebibliography}{10}

\bibitem{abutbul2020dnf}
Ami Abutbul, Gal Elidan, Liran Katzir, and Ran El-Yaniv, `Dnf-net: A neural
  architecture for tabular data', {\em arXiv preprint arXiv:2006.06465},
  (2020).

\bibitem{agarwal2020neural}
Rishabh Agarwal, Nicholas Frosst, Xuezhou Zhang, Rich Caruana, and Geoffrey~E
  Hinton, `Neural additive models: Interpretable machine learning with neural
  nets', {\em arXiv preprint arXiv:2004.13912}, (2020).

\bibitem{angelino2017learning}
Elaine Angelino, Nicholas Larus-Stone, Daniel Alabi, Margo Seltzer, and Cynthia
  Rudin, `Learning certifiably optimal rule lists for categorical data', {\em
  arXiv preprint arXiv:1704.01701}, (2017).

\bibitem{10.1007/978-3-031-25056-9_31}
Adrien Benamira, Thomas Peyrin, and Bryan~Hooi Kuen-Yew, `Truth-table net: A
  new convolutional architecture encodable by design into sat formulas', in
  {\em Computer Vision -- ECCV 2022 Workshops}, eds., Leonid Karlinsky, Tomer
  Michaeli, and Ko~Nishino, pp. 483--500, Cham, (2023). Springer Nature
  Switzerland.

\bibitem{bessiere2009minimising}
Christian Bessiere, Emmanuel Hebrard, and Barry O’Sullivan, `Minimising
  decision tree size as combinatorial optimisation', in {\em International
  Conference on Principles and Practice of Constraint Programming}, pp.
  173--187. Springer, (2009).

\bibitem{biere2009handbook}
Armin Biere, Marijn Heule, and Hans van Maaren, {\em Handbook of
  satisfiability}, volume 185, IOS press, 2009.

\bibitem{cohen1995fast}
William~W Cohen, `Fast effective rule induction', in {\em Machine learning
  proceedings 1995},  115--123, Elsevier, (1995).

\bibitem{cohen1999simple}
William~W Cohen and Yoram Singer, `A simple, fast, and effective rule learner',
  {\em AAAI/IAAI}, {\bf 99}(335-342), ~3, (1999).

\bibitem{dash2018boolean}
Sanjeeb Dash, Oktay Gunluk, and Dennis Wei, `Boolean decision rules via column
  generation', {\em Advances in neural information processing systems}, {\bf
  31}, (2018).

\bibitem{Dua:2019}
Dheeru Dua and Casey Graff.
\newblock {UCI} machine learning repository, 2017.

\bibitem{grewal2019application}
Jasleen~K Grewal, Basile Tessier-Cloutier, Martin Jones, Sitanshu Gakkhar,
  Yussanne Ma, Richard Moore, Andrew~J Mungall, Yongjun Zhao, Michael~D Taylor,
  Karen Gelmon, et~al., `Application of a neural network whole
  transcriptome--based pan-cancer method for diagnosis of primary and
  metastatic cancers', {\em JAMA network open}, {\bf 2}(4),  e192597--e192597,
  (2019).

\bibitem{ho1995random}
Tin~Kam Ho, `Random decision forests', in {\em Proceedings of 3rd international
  conference on document analysis and recognition}, volume~1, pp. 278--282.
  IEEE, (1995).

\bibitem{lakkaraju2016interpretable}
Himabindu Lakkaraju, Stephen~H Bach, and Jure Leskovec, `Interpretable decision
  sets: A joint framework for description and prediction', in {\em Proceedings
  of the 22nd ACM SIGKDD international conference on knowledge discovery and
  data mining}, pp. 1675--1684, (2016).

\bibitem{li2017comprehensive}
Yuanyuan Li, Kai Kang, Juno~M Krahn, Nicole Croutwater, Kevin Lee, David~M
  Umbach, and Leping Li, `A comprehensive genomic pan-cancer classification
  using the cancer genome atlas gene expression data', {\em BMC genomics}, {\bf
  18},  1--13, (2017).

\bibitem{liu2018integrated}
Jianfang Liu, Tara Lichtenberg, Katherine~A Hoadley, Laila~M Poisson,
  Alexander~J Lazar, Andrew~D Cherniack, Albert~J Kovatich, Christopher~C Benz,
  Douglas~A Levine, Adrian~V Lee, et~al., `An integrated tcga pan-cancer
  clinical data resource to drive high-quality survival outcome analytics',
  {\em Cell}, {\bf 173}(2),  400--416, (2018).

\bibitem{pedregosa2011scikit}
Fabian Pedregosa, Ga{\"e}l Varoquaux, Alexandre Gramfort, Vincent Michel,
  Bertrand Thirion, Olivier Grisel, Mathieu Blondel, Peter Prettenhofer, Ron
  Weiss, Vincent Dubourg, et~al., `Scikit-learn: Machine learning in python',
  {\em the Journal of machine Learning research}, {\bf 12},  2825--2830,
  (2011).

\bibitem{Puram2017SingleCell}
Siddhartha~V Puram, Itay Tirosh, Akash~S Parikh, Anoop~P Patel, and et~al.,
  `Single-cell transcriptomic analysis of primary and metastatic tumor
  ecosystems in head and neck cancer', {\em Cell}, {\bf 171}(7),
  1611--1624.e24, (Dec 2017).

\bibitem{quinlan2014c4}
J~Ross Quinlan, {\em C4. 5: programs for machine learning}, Elsevier, 2014.

\bibitem{revkov2023puree}
Egor Revkov, Tanmay Kulshrestha, Ken Wing-Kin Sung, and Anders~Jacobsen
  Skanderup, `Puree: accurate pan-cancer tumor purity estimation from gene
  expression data', {\em Communications Biology}, {\bf 6}(1),  394, (2023).

\bibitem{rivest1987learning}
Ronald~L Rivest, `Learning decision lists', {\em Machine learning}, {\bf 2}(3),
   229--246, (1987).

\bibitem{Tirosh2016Dissecting}
Itay Tirosh, Benjamin Izar, Sanjay~M Prakadan, Mark H~2nd Wadsworth, and
  et~al., `Dissecting the multicellular ecosystem of metastatic melanoma by
  single-cell rna-seq', {\em Science}, {\bf 352}(6282),  189--196, (Apr 2016).

\bibitem{tran2021fast}
Duc Tran, Hung Nguyen, Bang Tran, Carlo La~Vecchia, Hung~N Luu, and Tin Nguyen,
  `Fast and precise single-cell data analysis using a hierarchical
  autoencoder', {\em Nature communications}, {\bf 12}(1),  1029, (2021).

\bibitem{wang2021scalable}
Zhuo Wang, Wei Zhang, Ning Liu, and Jianyong Wang, `Scalable rule-based
  representation learning for interpretable classification', {\em Advances in
  Neural Information Processing Systems}, {\bf 34}, (2021).

\bibitem{wei2019generalized}
Dennis Wei, Sanjeeb Dash, Tian Gao, and Oktay Gunluk, `Generalized linear rule
  models', in {\em International Conference on Machine Learning}, pp.
  6687--6696. PMLR, (2019).

\bibitem{yang2021learning}
Fan Yang, Kai He, Linxiao Yang, Hongxia Du, Jingbang Yang, Bo~Yang, and Liang
  Sun, `Learning interpretable decision rule sets: A submodular optimization
  approach', {\em Advances in Neural Information Processing Systems}, {\bf 34},
  (2021).

\end{thebibliography}
\end{document}